\documentclass[a4paper,fleqn]{cas-dc}
\usepackage{amsmath,amsfonts}
\usepackage{algorithmic}
\usepackage{algorithm}
\usepackage{array}
\usepackage[caption=false,font=normalsize,labelfont=sf,textfont=sf]{subfig}
\usepackage{textcomp}
\usepackage{stfloats}
\usepackage{url}
\usepackage{verbatim}
\usepackage{graphicx}
\usepackage{cite}
\usepackage{soul,xcolor}
\usepackage{booktabs}
\usepackage{multirow}
\usepackage{threeparttable}
\usepackage{hyperref}
\hypersetup{hidelinks}        
\usepackage{xcolor}
\usepackage[switch]{lineno}
\usepackage{graphicx}
\usepackage{bm} 

\usepackage{tabularx,array,booktabs}       
\usepackage{colortbl,xcolor}
\usepackage{multirow,makecell}
\usepackage{tabularx} 

\newcolumntype{A}{>{\raggedright\arraybackslash}m{1cm}} 
\newcolumntype{B}{>{\centering\arraybackslash}m{1.38cm}} 
\newcolumntype{Y}{>{\centering\arraybackslash}X}   

\usepackage[normalem]{ulem}

\usepackage[numbers]{natbib}

\def\tsc#1{\csdef{#1}{\textsc{\lowercase{#1}}\xspace}}
\tsc{WGM}
\tsc{QE}
\tsc{EP}
\tsc{PMS}
\tsc{BEC}
\tsc{DE}

\begin{document}
\let\WriteBookmarks\relax
\def\floatpagepagefraction{1}
\def\textpagefraction{.001}
\shorttitle{High-Energy Concentration for Federated Learning in Frequency Domain}
\shortauthors{Weiying Xie et~al.}

\title [mode = title]{High-Energy Concentration for Federated Learning in Frequency Domain}

\author[1]{Weiying\ Xie}
\ead{wyxie@xidian.edu.cn}
\credit{Conceptualization, Supervision}

\author[1]{Haozhi\ Shi}[orcid=0000-0002-3733-507X]
\cormark[1]
\ead{shz@stu.xidian.edu.cn}
\credit{Methodology, Experiments, Writing--original draft}

\author[1]{Hangyu\ Ye}
\ead{wagyr@stu.xidian.edu.cn}
\credit{Validation, Software}

\author[1]{Daixun\ Li}
\ead{ldx@stu.xidian.edu.cn}
\credit{Validation, Visualization}

\author[1]{Jitao\ Ma}
\ead{21011210271@stu.xidian.edu.cn}
\credit{Data curation}

\author[1]{Yunsong\ Li}
\ead{ysli@mail.xidian.edu.cn}
\credit{Review, Resources}

\author[2]{Leyuan\ Fang}
\ead{fangleyuan@gmail.com}
\credit{Review, Editing}

\affiliation[1]{organization={State Key Laboratory of Integrated Services Networks, Xidian University},
                addressline={},
                city={Xi'an},
                state={Shaanxi},
                postcode={710071},
                country={China}}

\affiliation[2]{organization={College of Electrical and Information Engineering, Hunan University},
                addressline={},
                city={Changsha},
                state={Hunan},
                postcode={410082},
                country={China}}

\cortext[1]{Corresponding author: Haozhi Shi (shz@stu.xidian.edu.cn).}

\fntext[fund1]{This work was supported in part by the National Natural Science Foundation of China under Grants 62322117, 62371365, U24B20136, and U22B2014, and in part by the Fundamental Research Funds for the Central Universities under Grant ZYTS24101.}


\begin{abstract}
Federated learning (FL) facilitates the mitigation of data heterogeneity by transmitting privacy-preserving synthetic data. However, existing aggregation-free frameworks are limited by the need to optimize and transmit the entire spatial domain, resulting in redundant information and noise that impact global model performance and increase communication burden. To alleviate this issue, we propose a novel Frequency-Domain–aware FL method (FedFD), inspired by the energy concentration and component orthogonality of the frequency domain. The principle behind FedFD is that low-energy like high-frequency components usually contain redundant information and noise, thus filtering them helps reduce communication costs and optimize performance. Specifically, the discrete cosine transform is first used to convert spatial domain features into the frequency domain. Subsequently, the low-frequency components are optimized via a designed frequency-domain alignment objective, which differs from existing spatial domain alignment. On this basis, real data-driven synthetic classification is introduced to enhance the quality of low-frequency components. Furthermore, a low-frequency-based curriculum-style communication strategy is designed to further enhance global model performance by gradually increasing the low-frequency component increments while maintaining a low communication budget. On five image and speech datasets, FedFD achieves superior performance than state-of-the-art methods while reducing communication costs. For example, on the CIFAR-10 with Dirichlet coefficient $\alpha = 0.01$, FedFD achieves a minimum reduction of 37.78\% in the communication cost, while attaining a 10.88\% performance gain.
\end{abstract}




\maketitle

\section{Introduction}

\begin{figure}
  \centering
  \includegraphics[width=\linewidth]{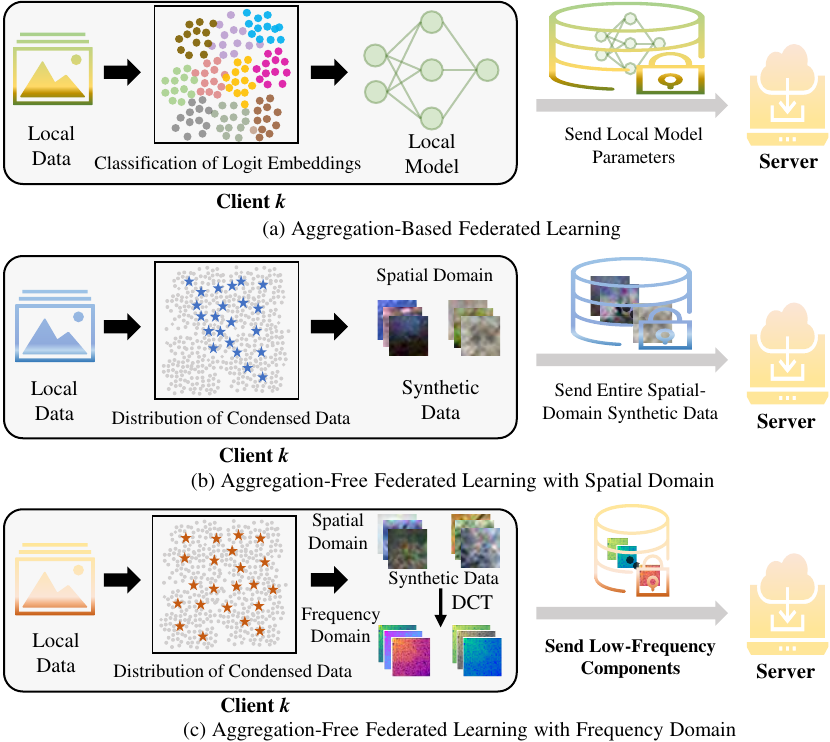}
  \caption{Illustration of different federated learning frameworks. (a) Sending optimized local model parameters. (b) Sending entire spatial-domain synthetic data. (c) Our FedFD: Sending low-frequency components of synthetic data.}
  \label{Fig1}
\end{figure}

Federated Learning (FL) \cite{2}\cite{56}\cite{57}, as a distributed training paradigm with privacy protection, has received widespread attention in multimedia fields \cite{55}\cite{3}\cite{51}. As illustrated in Fig. \ref{Fig1}(a), typical FL methods adopt an aggregation-based framework \cite{2}\cite{5}\cite{53}, aggregating local models from each client to update the global model on the server. Most existing efforts address data heterogeneity through two primary approaches \cite{47}. One approach introduces regularization terms on the client to prevent local models from deviating excessively from the global consensus \cite{6}\cite{8}, while another designs specific aggregation frameworks on the server for adjusting the contribution of the local model to the global model \cite{10}. Nevertheless, such aggregation-based FL methods still have limitations in terms of robustness and communication costs, especially when facing severe data heterogeneity \cite{12}\cite{13} and using more complex models \cite{14}\cite{48}. Our subsequent experiments also validate this point.

DD learns a small set of synthetic samples to approximate the training performance of the full dataset \cite{17}. Within the DD paradigm, gradient matching \cite{22}\cite{23} and trajectory matching \cite{24}\cite{25} are relatively effective, as they align the optimization dynamics between real and synthetic data. Nevertheless, their reliance on bi-level optimization introduces substantial computational overhead and optimization difficulties \cite{26}. Moreover, trajectory matching significantly increases memory usage, limiting scalability in practical deployments \cite{27}. By comparison, Distribution Matching (DM) \cite{28} directly aligns feature distributions in the embedding space. It achieves a better balance between efficiency and performance, offering a more practical pathway for DD \cite{29}\cite{30}.

Benefiting from dataset distillation \cite{27}\cite{28}, recent research focuses on alleviating the above phenomenon by sending synthetic data to the server, which is called the aggregation-free method \cite{15}\cite{16}\cite{17}. As shown in Fig. \ref{Fig1}(b), each client captures the distribution of the real data set by generating synthetic data and then sends it to the server for joint training of the global model. However, the limitation of comprehensive efficiency is inevitable in such federated settings due to (1) the design of sending entire spatial domain and (2) the objective of single-view distribution matching (DM) \cite{18}. On the one hand, given the difficulty of distinguishing regional criticality in the spatial domain, sending the entire synthetic data is required \cite{19}, which increases the communication burden due to the redundant information and noise it carries. On the other hand, the single-view DM objective weakens the quality of synthetic data by failing to exploit inter-class discrimination, thereby leading to suboptimal performance of the global model. Overall, this reveals that there is still significant potential for improvement in both communication efficiency and model performance.

In light of the above analysis, this paper presents a high-energy concentration method for federated learning in the frequency domain, named FedFD, which effectively utilizes the energy distribution and component orthogonality of the discrete cosine transform (DCT). As illustrated in Fig. \ref{Fig1}(c), the key idea of FedFD is to promote the quality of low-frequency components while filtering out high-frequency components containing noise. Specifically, frequency-domain-assisted dual-view coordination is designed to optimize the low-frequency components of synthetic data. Among them, frequency-domain distribution alignment enhances the quality of low-frequency components by calculating the distribution loss in the frequency domain. On this basis, inter-class discrimination is improved by the real data-driven synthetic classification objective. Subsequently, we introduce a low-frequency-based curriculum-style communication strategy. In this way, the number of low-frequency components sent gradually increases, thus enhancing the representation capability of the global model. Extensive experiments demonstrate that the proposed FedFD achieves superior performance while maintaining high communication efficiency. The contributions of this work can be summarized as follows:
\begin{itemize}
\item We propose a novel high-energy concentration method for FL (FedFD), which is the first to explore the potential of aggregation-free FL in the frequency domain from the perspective of energy distribution. 
\item We design a frequency-domain assisted dual-view coordination objective to optimize the low-frequency components that carry most energy in the synthetic data. Unlike spatial alignment, this objective not only promotes the alignment of low-frequency components by transferring to the frequency domain, but also enhances the quality by improving the inter-class discrimination.
\item We present a low-frequency-based curriculum-style communication strategy regarding for FL to enhance the representation capacity of the global model with lower communication costs.
\item We conduct extensive experiments on five datasets from different modalities, including image classification and speech command classification. The results demonstrate that FedFD outperforms state-of-the-art FL methods in both performance and communication efficiency.
\end{itemize}

\begin{figure*}
    \begin{center}
        \includegraphics[width=1.0\linewidth]{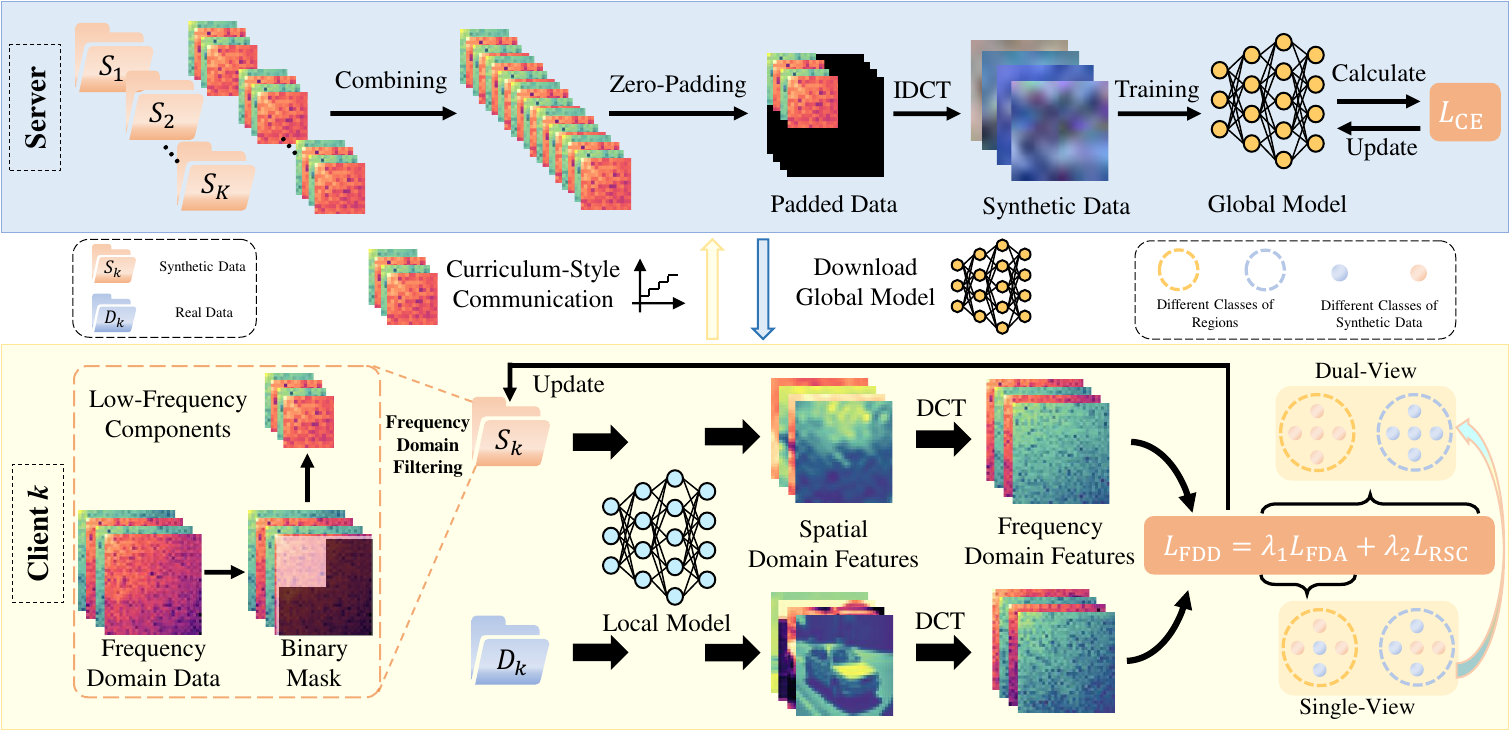}
        \caption{Overall architecture of FedFD. The lower left corner illustrates the key of FedFD: retaining low-frequency components with high-energy concentration via frequency domain filtering. Notably, a frequency-domain-assisted dual-view coordination loss $\mathcal{L}_\mathrm{FDD}$ is designed to optimize the low-frequency components of $S_k$, as shown  at the bottom. Subsequently, a low-frequency-based curriculum-style communication strategy is adopted to enhance the performance of the global model. Finally, on the server, the low-frequency data is zero-padded and processed using the IDCT to train the global model.}
        \label{Fig2}
    \end{center}
\end{figure*}

\section{Related work}
\subsection{Federated learning}
Federated learning system focuses on the collaborative modeling of distributed data and has won great attention in multimedia fields such as image diagnosis and voice interaction in recent years \cite{3}. A substantial amount of research has been devoted to exploring optimization strategies for Non-Independent and Identically Distributed (Non-IID) data \cite{20}\cite{21}\cite{22}. As the pioneering algorithm for FL, FedAvg \cite{2} establishes the basis for collaborative modeling of distributed data by adopting a simple aggregation strategy. Li et al. \cite{5} proposed FedProx, which introduces a proximal regularization term to constrain the deviation between the local and global models. FedNova \cite{22} was designed to alleviate inconsistencies in local update objectives across clients by employing a normalized averaging strategy. Karimireddy et al. \cite{6} developed SCAFFOLD, which corrects local update biases through the introduction of control variates. These aggregation-based methods demonstrate a progressive evolution from simple averaging to refined optimization strategies for tackling data heterogeneity challenges. Nevertheless, when dealing with different levels of data heterogeneity and more complex models, these methods still have considerable potential for improvement in terms of generalization ability and communication efficiency \cite{16}. 

Recently, aggregation-free methods have gradually attracted research interest due to their effectiveness in addressing the aforementioned challenges \cite{17}. FedDM \cite{15} was proposed to reduce the discrepancy between synthetic and real data by introducing a distribution matching loss \cite{18}. Subsequently, the global model is trained using synthetic data from all clients. Building upon this, Wang et al. \cite{16} designed FedAF to further improve the performance of the global model by incorporating a regularization term. Although these two studies have made significant progress, they primarily generate synthetic data based on DM loss, which results in a lack of inter-class discrimination compared to real data. Moreover, these methods focus solely on spatial-domain data, leading to limitations in their performance and efficiency.
\subsection{Dataset distillation}
Early dataset distillation algorithms typically adopt a bi-level optimization strategy. For example, Zhao et al. \cite{24} proposed condensing data by matching the gradients of real data and synthetic data. However, this method necessitates model updates during optimization, leading to large computational overhead. To mitigate this burden, DM \cite{18} was proposed to generate synthetic data through simple and effective distribution matching. Shin et al. \cite{19} proposed FreD, which optimizes frequency-domain information by computing losses in the spatial domain. It then selectively retains frequency-domain components based on the Explained Variance Ratio (EVR). Under the same budget, the algorithm improves the performance by increasing the increment, which reflects the potential of frequency domain in dataset distillation. Nevertheless, FreD exhibits certain limitations when applied to FL: 1) updating frequency-domain information via spatial-domain losses makes it difficult to directly align high and low-frequency components; 2) EVR-based component selection introduces additional computational and communication cost. It is evident that the goal of dataset distillation is to achieve data lightweighting \cite{26}\cite{27}, which aligns well with the current requirements of aggregation-free FL. Consequently, a communication-efficient and degradation-free theme may drive the development of FL.

\section{Methodology}
\subsection{Preliminaries}
\begin{algorithm}[t]
\raggedright
\caption{FedFD}
\label{alg:FedFD}
\begin{algorithmic}[1]
\STATE \textbf{Input:} Real data set $D$; Initialized synthetic data set $S$; Initialized network parameters $\mathrm{w}$; Number of local iterations $T$; Learning rates $\eta_l$ and $\eta_g$; DCT $F(\cdot)$.\\
\STATE \textbf{Output:} Optimal global model $\mathrm{w^*}$ .\\
\STATE \textbf{Server executes:}

\FOR{each round $r = 1, \dots, R$}
    \FOR{each client $k = 1, \dots, K$}
        \STATE $S_k^{fre} \leftarrow \text{ClientUpdate}(k, \mathrm{w_r})$
        \STATE Pad with zeros and perform IDCT operation: $S_k \leftarrow F^{-1}(ZeroPad(S_k^{fre}))$
        \STATE Send $S_k$ to the server
    \ENDFOR
    \STATE Combine data from all clients: $S = \{S_k\}_{k=1}^{K}$
    \STATE Update weights: $\mathrm{w_{r+1}} \gets \mathrm{w_r} - \eta_g \nabla_\mathrm{w} \mathcal{L}_{CE}(\mathrm{w_r}, S)$
\ENDFOR
\STATE \textbf{ClientUpdate}$(k, \mathrm{w_r})$:
\FOR{$t = 0, \dots, T-1$}
    \STATE Sample mini-batch pairs $(B_c^{D_k}, B_c^{S_k})$ from $D_k$ and $S_k$ for each class $c$
    \STATE Compute $\mathcal{L}_\mathrm{FDD}$ by Eq. (\ref{11}): $\mathcal{L}_\mathrm{FDD} \leftarrow \sum_{c=0}^{C-1}{\mathcal{L}_\mathrm{FDD}^c}(F(B_c^{D_k}), F(B_c^{S_k}))$
    \STATE Update synthetic data: $S_k \leftarrow S_k - \eta_l \nabla_{S_k}\mathcal{L}_\mathrm{FDD}$
    \STATE Filter in frequency domain: $S_k^{fre} \leftarrow M_{x, y}\odot F(S_k)$
\ENDFOR
\end{algorithmic}
\end{algorithm}
The raw data set collected from \( K \) clients is denoted as \( D = \{D_k\}_{k=1}^{K} \), where \( D_k \) represents the real data on client \( k \), and \( |\cdot| \) indicates the number of samples in the corresponding client. The typical objective of aggregation-based FL is to update a global model through collaborative optimization across clients, which is formulated as:
\begin{equation}\arg\min_{\mathrm{w}}\mathcal{L}_{g}(\mathrm{w})=\sum_{k=1}^{K}p_{k}\mathcal{L}_{k}(\mathrm{w},D_{k}),\end{equation}
where $\mathrm{w}$ represents the optimizable parameter, and $p_k$ denotes the contribution of client $k$, satisfying $\sum_{k=1}^Kp_k=1$. $\mathcal{L}_g(\cdot)$ and $\mathcal{L}_k(\cdot)$ represent the objective functions of the server and client $k$, respectively. Due to data heterogeneity among clients, the server finds it challenging to capture the optimal update direction through the interaction between clients \cite{15}. Therefore, the aggregation-free FL method is proposed, aiming to collaboratively optimize the global model using the synthetic data \( S = \{S_k\}_{k=1}^{K} \) from all clients.

The problem of optimizing the synthetic data $S_k$ at client $k$ is formulated as follows:
\begin{equation}\arg\min_{S_k}\mathcal{L}_{\mathrm{DM}}(S_k,D_k)=\sum_{c=0}^{C_k-1}\|\mu_{k,c}^{S}-\mu_{k,c}^{D}\|^2,\end{equation}
where $\mathcal{L}_\mathrm{DM}(\cdot)$ describes the objective function of distribution matching, $C_k$ indicates the total number of classes in client $k$. $\mu_{k,c}^{S}$ and $\mu_{k,c}^{D}$ denote feature means, which are expressed as:
\begin{equation}\mu_{k,c}^{S}=\frac{1}{|S_{k}^c|}\sum_{i=1}^{|S_{k}^c|}E_{\mathrm{w}}(s_{k,c}^i),
\mu_{k,c}^{D}=\frac{1}{|D_{k}^c|}\sum_{i=1}^{|D_{k}^c|}E_{\mathrm{w}}(d_{k,c}^i),
\end{equation}
where $E_{\mathrm{w}}(\cdot)$ represents the feature extractor, which is a local model without the last classification layer. $d_{k,c}^i\in{D_k}$ and $s_{k,c}^i\in{S_k}$ are the $i$-th sample of class c, respectively.

Aggregation-free FL has emerged to overcome the inherent limitations of aggregation-based architectures when dealing with data heterogeneity \cite{17}. Through Theorems 1 and 2, we offer a simple yet intuitive analysis of the advantages of FL without aggregation.

\textbf{Theorem 1.} The relationship between the global model $W^*$ obtained by centralized training and the global model $W_g$ in aggregation-based FL is as follows:
\begin{equation}W_g=\sum_{k=1}^Kp_kW_k=p_k(W^*+\Delta_k)=W^*+\sum_{k=1}^Kp_k\Delta_k,\label{4}\end{equation}
where $\Delta_k$ indicates the deviation caused by data heterogeneity. In contrast, the optimization problem for aggregation-free FL is expressed as follows:
\begin{equation}
W_g^{\prime}=\arg\min_\mathrm{w}\mathcal{L}_g(\mathrm{w},S).\label{5}
\end{equation}
Eq. (\ref{4}) intuitively shows that due to the impact of data heterogeneity, the global model \(W_g\) maintains a certain distance from the optimal model \(W^*\), which is given by \(\sum_{k=1}^{K} p_k\Delta_k\). This suggests that aggregation-based frameworks are inherently susceptible to data heterogeneity. In contrast, the synthetic data \(S\) in Eq. (\ref{5}) is less affected by data heterogeneity, as it is primarily determined by the quality of the client-generated synthetic data \(S_k\), where \(S = \{S_k\}_{k=1}^{K}\). When $S_k$ approximating $D_k$, we obtain $S$ approximating $D$, leading the global model \(W_g'\) to converge closely to the optimal solution \(W^*\). This demonstrates that, compared to aggregation-based FL, aggregation-free FL exhibits stronger robustness in the presence of data heterogeneity. Subsequently, Theorem 2 illustrates the feasibility of $S_k$ approximating $D_k$.

Assumption: The objective function \( \mathcal{L}_k^\prime(\theta) \) is strongly convex with respect to \( \theta \) and is $\mathcal{L}$-$smooth$, ensuring the existence of a global optimum \( \theta^\ast \).

\textbf{Theorem 2.} 
The synthetic data is treated as an updatable parameter \(\theta\). The distance between $\theta_{t+1}$ and $\theta^*$ is first calculated:
\begin{equation}\begin{aligned}
\parallel\theta_{t+1}-\theta^*\parallel^2 & =\parallel\theta_t-\eta\nabla \mathcal{L}_k^\prime(\theta_t)-\theta^*\parallel^2 \\
 & =\parallel\theta_t-\theta^*\parallel^2-2\eta\nabla \mathcal{L}_k^\prime(\theta_t)^\top(\theta_t-\theta^*) \\
& \quad +\eta^2\parallel\nabla \mathcal{L}_k^\prime(\theta_t)\parallel^2.
\end{aligned}\end{equation}
\(\theta^*\) is the global optimum of \(\mathcal{L}_k^\prime(\theta)\), meaning that:
\begin{equation}\mathcal{L}_k^\prime(\theta_t)\geq \mathcal{L}_k^\prime(\theta^*). \end{equation}
Based on the properties of \(\mathcal{L}_k^\prime(\theta)\), for any \(\theta\) and \(x\), we have:
\begin{equation}\mathcal{L}_k^\prime(x)\geq \mathcal{L}_k^\prime(\theta)+\nabla \mathcal{L}_k^\prime(\theta)^\top(x-\theta)+\frac{\alpha}{2}\|x-\theta\|^2,\end{equation}
where $\alpha>0$. Let $x=\theta^*,\theta=\theta_t$, then:
\begin{equation}\mathcal{L}_k^\prime(\theta^*)\geq \mathcal{L}_k^\prime(\theta_t)+\nabla \mathcal{L}_k^\prime(\theta_t)^\top(\theta^*-\theta_t)+\frac{\alpha}{2}\|\theta^*-\theta_t\|^2.\end{equation}
Combining Eqs. (2) and (4), we get:
\begin{equation}\mathcal{L}_k^\prime(\theta_t)\geq \mathcal{L}_k^\prime(\theta^*)\geq \mathcal{L}_k^\prime(\theta_t)+\nabla \mathcal{L}_k^\prime(\theta_t)^\top(\theta^*-\theta_t)+\frac{\alpha}{2}\|\theta^*-\theta_t\|^2.\end{equation}
Then we have:
\begin{equation}\nabla \mathcal{L}_k^\prime(\theta_t)^\top(\theta_t-\theta^*)\geq\frac{\alpha}{2}\|\theta^*-\theta_t\|^2=\frac{\alpha}{2}\|\theta_t-\theta^*\|^2.\end{equation}
It can be known from $L$-$smooth$:
\begin{equation}\|\nabla \mathcal{L}_k^\prime(\theta_t)-\nabla \mathcal{L}_k^\prime(\theta^*)\|\leqslant\beta\|\theta_t-\theta^*\|,\end{equation}
where the smoothness coefficient is denoted by $\beta > 0$. Additionally, at the optimal point \(\theta^*\):
\begin{equation}\nabla \mathcal{L}_k^\prime(\theta^*)=0.\end{equation}
So:
\begin{equation}\|\nabla \mathcal{L}_k^\prime(\theta_t)\|^2\leq\beta^2\|\theta_t-\theta^*\|^2.\end{equation}
Substituting Eq. (13) into Eq. (6) yields:
\begin{equation}
\begin{aligned}
\parallel\theta_{t+1}-\theta^*\parallel^2 & \leq \parallel\theta_t-\theta^*\parallel^2 - 2\eta  \frac{\alpha}{2} \parallel\theta_t-\theta^*\parallel^2 \\
& \quad + \eta^2 \beta^2 \parallel\theta_t-\theta^*\parallel^2,
\end{aligned}
\end{equation}
where \(\eta = \frac{\alpha}{2\beta^2}\). The synthetic data is treated as an optimizable parameter \(\theta\), and the relationship between iterations is expressed as follows:
\begin{equation}
\begin{aligned}
\parallel\theta_{t+1}-\theta^*\parallel^2 & \leq \parallel\theta_t-\theta^*\parallel^2 - 2\eta  \frac{\alpha}{2} \parallel\theta_t-\theta^*\parallel^2 \\
& \quad + \eta^2 \beta^2 \parallel\theta_t-\theta^*\parallel^2 \\
& \leq \kappa \parallel\theta_t-\theta^*\parallel^2.
\end{aligned}
\label{6}
\end{equation}
where \(\kappa = 1 - \eta\alpha + \eta^2\beta^2\). Thus we can get the following:
\begin{equation}\parallel\theta_T-\theta^*\parallel^2\leq\kappa^T\parallel\theta_0-\theta^*\parallel^2.\end{equation}
It is also known that \( 0 < \kappa < 1 \), and as the number of iterations \( T \) becomes sufficiently large, \( \theta_T \rightarrow \theta^\ast \). Since \( \mathcal{L}_k^\prime(\theta) \) is the loss function that measures the distributional difference between synthetic and real data, it can be assumed that at \( \theta^\ast \), \( S_k \approx D_k \). Therefore, as the number of iterations increases, it becomes feasible for \( S_k \) to approximate \( D_k \).

The analysis explores the performance of aggregation-based and aggregation-free FL when faced with different Non-IID data distributions. This is further validated in the subsequent experimental section.

\textbf{Theorem 3.}
We characterize communication security primarily by the likelihood that an attacker can successfully extract the original data from the sent synthetic data \cite{51}\cite{19}. Using the entire spatial-domain data $S$, we adopt a maximum likelihood estimation approach to estimate its recovery distribution:
\begin{equation}\hat{D}=\arg\max_{D^{\prime}}P(S|D^{\prime}),\end{equation}
where \(\hat{D}\) is the distribution deduced by the attacker that is most likely to approximate the real data, and \(D^\prime\) represents all possible distributions. For the partial frequency-domain data \(F\), it is expressed as:
\begin{equation}\hat{D}=\arg\max_{D^{\prime}}P(F|D^{\prime}).\end{equation}
Due to the binary mask matrix consistently selecting information from the top-left corner, our method can omit the position index matrix when sending the sparse matrix. Consequently, an attacker must guess the combination of position indices for each sample in \(F\). The number of combinations is:
\begin{equation}\begin{aligned}
C(d\times d,l)\times l! & =\frac{(d\times d)!}{l!\cdot(d\times d-l)!}\times l! \\
 & =\frac{(d\times d)!}{(d\times d-l)!},
\end{aligned}\end{equation}
where $d$ represents the edge length of the entire synthetic data, and the total length of the filtered synthetic data is denoted as $l=s\times s$. Notably, it is necessary to pad \( d \times d - l \) frequency components. The reconstruction error is given by:
\begin{equation}\mathbb{E}\left\|F_p-F_p^{\prime}\right\|_2^2\propto k\cdot(d\times d-D)\cdot\sigma_f^2,\end{equation}
where \( F \) and \( F' \) represent the real and predicted frequency components, respectively, \( \sigma_f^2 \) denotes the variance of the frequency components, and $k$ is a positive adjustment coefficient. This means that the information entropy of the sparse frequency-domain data is higher, indicating greater randomness. Therefore, based on Eqs. (18) and (19), it can be inferred that the sparse frequency-domain communication strategy has a lower attack success probability compared to the entire spatial-domain communication strategy.

\begin{figure}
  \centering
  \includegraphics[width=\linewidth]{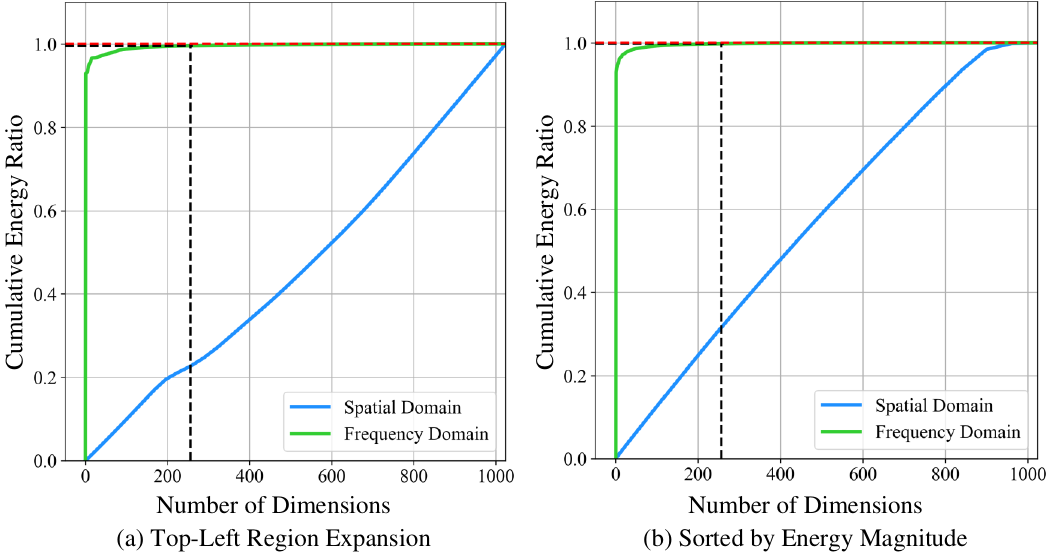}
  \caption{Comparison of cumulative energy ratios in the spatial and frequency domains as dimensions increase. (a) Dimensions are added sequentially from the top-left corner. (b) Dimensions are added in descending order of energy.}
  \label{Fig3}
\end{figure}

\subsection{Motivation}
Compared to the spatial domain, the frequency domain concentrates the energy in a specific region with only a few dimensions. To investigate whether the energy concentration is higher in the frequency domain, we compare how energy changes with the number of dimensions in both the spatial and frequency domains. As shown in Figs. \ref{Fig3}(a) and (b), only a few components in the frequency domain are needed for the cumulative energy ratio to tend to 1, whether the number of dimensions increases sequentially from the top-left corner or by increasing the number of dimensions based on energy ranking from high to low. This is primarily because the energy in the frequency domain is mostly concentrated in the low-frequency components \cite{29}. Therefore, the advantage of using the frequency domain in FL is that it reduces storage requirements by retaining only a few high-energy components, thus improving communication efficiency.

In this paper, Eq. (5) is adjusted as follows:
\begin{equation}W_g^{\prime}=\arg\min_\mathrm{w}\mathcal{L}_g(\mathrm{w},F^{-1}(ZeroPad(S_k^{fre}))),\label{5a}\end{equation}
where \(ZeroPad(\cdot)\) denotes the zero-padding operation, and \( F^{-1}(\cdot) \) represents the inverse discrete cosine transform (IDCT). Based on Eq. (8), the key to FedFD lies in acquiring and optimizing \( S_k^{fre} \). As shown in Fig. \ref{Fig3}, although the energy enhancement trends are similar for both methods, the explicit positional information in energy ranking increases storage requirements. Furthermore, considering that the energy in the DCT-based spectrogram is mainly concentrated in the top-left corner. Therefore, before sending to the server, we perform frequency domain filtering using binary masks on the client to capture the low-frequency components. The specific operation is illustrated in the lower-left corner of Fig. \ref{Fig2} and can be formulated as:
\begin{equation}
S_k^{fre} = M_{x,y} \odot f_k = 
\begin{cases} 
f_{k}^{x,y} & \text{if } M^{x,y} = 1 \\ 
0 & \text{otherwise}
\end{cases},
\end{equation}
where $f_k$ and $M^{x,y}$ represent the frequency-domain features and binary mask, respectively. For ease of operation, we typically set the values within a square window of size $s$ in the top-left corner to 1, while the remaining values are set to 0. The pipeline of FedFD is outlined in Algorithm \ref{alg:FedFD}.

\subsection{Frequency-domain-assisted dual-view coordination}
Research shows that simple and effective distribution matching scheme enables synthetic data to learn the distribution of real data \cite{15}\cite{18}. As shown in the upper right of Fig. \ref{Fig2}, in a single-view setting, although distribution alignment is achieved, classification alignment is still limited, leading to weak inter-class discrimination. Therefore, adopting a dual-view approach helps enhance the consistency between synthetic and real data at a deeper level. Furthermore, to strengthen the alignment of low-frequency components, we design a frequency-domain-assisted dual-view coordination objective, which includes frequency-domain distribution alignment and real data-driven synthetic classification.

\textbf{Frequency-domain distribution alignment.} Inspired by the concept of simple and efficient distribution consistency and the advantages of the frequency domain in FL, as detailed in the following section, we explore a frequency-domain distribution alignment approach, as follows:
\begin{equation}\mathcal{L}_{\mathrm{FDA}}(S_k,D_k)=\sum_{c=0}^{C_k-1}\parallel\nu_{k,c}^S-\nu_{k,c}^D\parallel^2,\end{equation}
where $\nu_{k,c}^S$ and $\nu_{k,c}^D$ represent the averages of the frequency-domain feature representations, expressed as:
\begin{equation}\nu_{k,c}^S=\frac{1}{|S_k^c|}\sum_{i=1}^{|S_k^c|}F(E_{\mathrm{w}}(s_{k,c}^i)),\nu_{k,c}^D=\frac{1}{|D_k^c|}\sum_{i=1}^{|D_k^c|}F(E_{\mathrm{w}}(d_{k,c}^i)),\end{equation}
where $F(\cdot)$ indicates the DCT operation. Compared to $\mathcal{L}_\mathrm{DM}$, $\mathcal{L}_\mathrm{FDA}$ measures the distribution difference of low-frequency components more directly and effectively.

\textbf{Real data-driven synthetic classification.} 
Considering only distribution alignment may lack the full exploitation of the discrimination of synthetic data. To address this, the common approach is to directly introduce the classification loss of synthetic data. However, the regularization term overlooks the classification performance of real data \cite{25}\cite{30}. Optimizing only the classification loss of synthetic data without considering its consistency with real data may lead to suboptimal distillation effects. Based on this, we enhance the discrimination of synthetic data by guiding it with the classification performance of real data, expressed as follows:
\begin{equation}\mathcal{L}_{\mathrm{RSC}}=\mathcal{L}_\mathrm{CE}(S_k)\cdot\left(\frac{1}{|D_k|}\sum_{i=1}^{|D_k|}\mathbb{I}(p_i=l_i)\right),\end{equation}
where \( p_i \) and \( l_i \) represent the predicted result and the corresponding label of the \( i \)-th sample, respectively. \( \mathcal{L}_{\text{CE}}(\cdot) \) denotes the cross-entropy loss. \( \mathbb{I}(\cdot) \) is an indicator function that equals 1 when the condition is satisfied and 0 otherwise.

In summary, the frequency-domain-assisted dual-view coordination objective function $\mathcal{L}_\mathrm{FDD}$ is given as follows:
\begin{equation}\mathcal{L}_\mathrm{FDD}=\lambda_1\mathcal{L}_\mathrm{FDA}+\lambda_2\mathcal{L}_\mathrm{RSC}, \label{11} \end{equation}
where $\lambda_1$ and $\lambda_2$ are hyperparameters used to balance the different terms. The goal is for the synthetic data not to directly mimic the original distribution of real data, but to maintain consistency in the frequency-domain dimension. In addition, deeper feature representations are more abstract than shallow ones, meaning privacy-sensitive information may have been blurred during the alignment process \cite{31}. At the same time, average alignment results in a non-one-to-one correspondence between synthetic and real data, which reduces reliance on individual real samples \cite{49}. Based on the above analysis, it is clear that the designed objective holds significant potential for privacy protection. Correspondingly, we also perform further validation through visualization.

\begin{figure}
  \centering
  \includegraphics[width=0.7\linewidth]{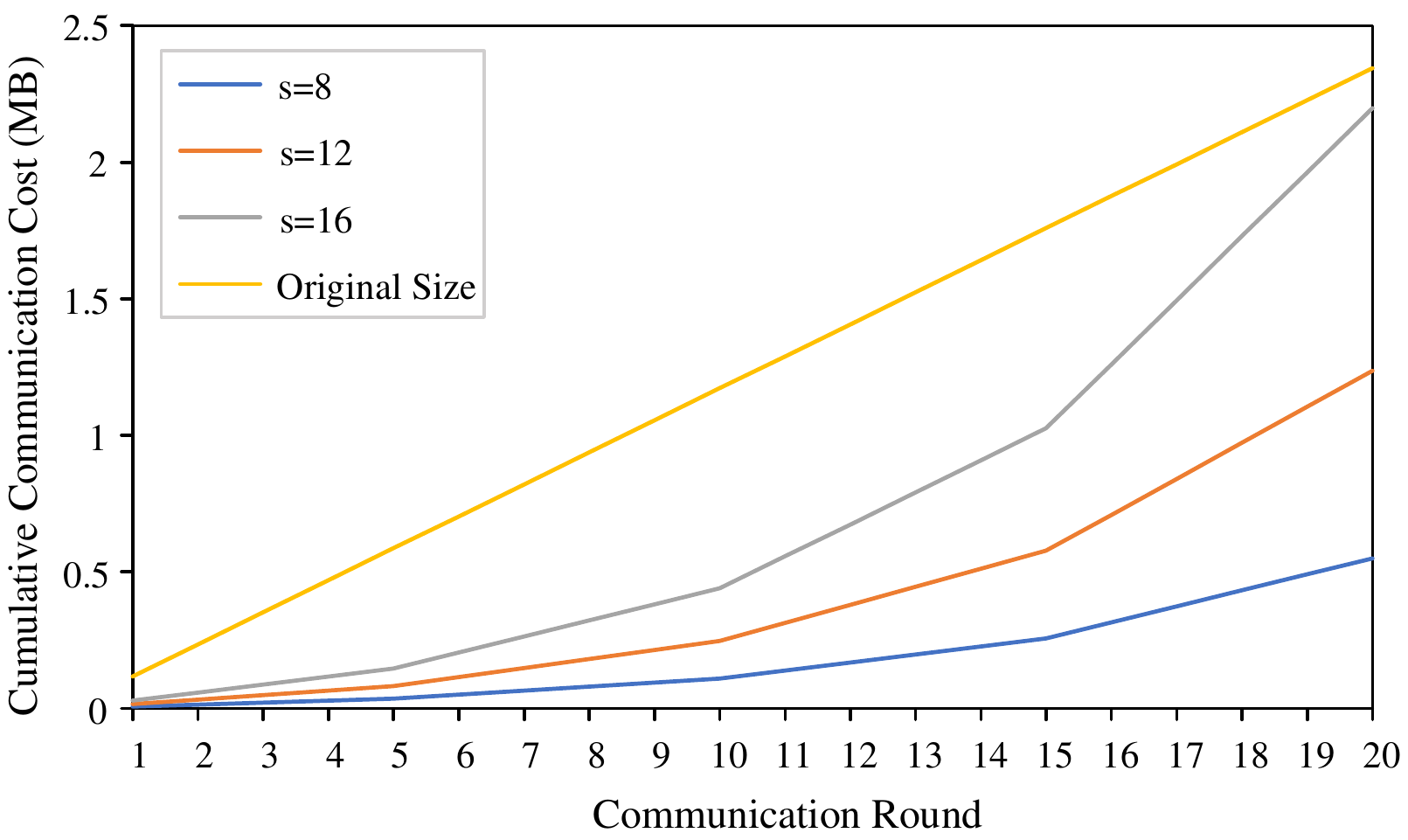}
  \caption{Comparison of the cumulative communication costs of spatial and frequency domain communication strategies.}
  \label{Fig8}
\end{figure}

\subsection{Low-frequency-based curriculum-style communication}
Studies show that increasing instance increments under the same storage budget contributes to improved performance \cite{33}\cite{34}. To this end, a low-frequency-based curriculum-style communication strategy is designed, aiming to improve communication efficiency while ensuring no performance degradation. Specifically, the low-frequency-based curriculum-style communication strategy is employed to send synthetic data. This strategy evenly divides the entire communication process into four stages, where the instance increment follows a linear growth pattern with a coefficient of $m$, making it progressively easier to train the global model on the server. The global model parameter update expression is as follows:
\begin{equation}\mathrm{w}_{t+1}=\mathrm{w}_t-\eta\nabla_\mathrm{w} \mathcal{L}(\mathrm{w}_t),\end{equation}
where $\mathcal{L}(\mathrm{w}_t)$ is expanded as follows:
\begin{equation}\mathcal{L}(\mathrm{w}_t)=\frac{1}{N}\sum_{k=1}^K\sum_{c=0}^{C_k-1}\sum_{n=1}^{N_{k,c}}\mathcal{L}_{\mathrm{CE}}(\mathrm{w}_t),\end{equation}
where $N$ is the total number of synthetic data samples.
It can be seen that according to the curriculum strategy, the value of $N_{k,c}$ is as follows:
\begin{equation}N_{k,c}=m\cdot (g-1)+b,\end{equation}
where $g$ and $b$ are the number of stages and the initial value of the first stage, respectively. At each stage, the sent frequency-domain components always require less storage than the entire spatial-domain data. To facilitate a more intuitive comparison, we present the variation in cumulative communication cost as the communication rounds progress. 

\begin{table*}
  \centering
  \caption{Performance ($\bm{\%}$) comparison of FL methods with different frameworks on CIFAR-10, CIFAR-100, and SVHN datasets at three levels of heterogeneity ($\bm{\alpha}$).}
  \label{tab1}
  \renewcommand{\arraystretch}{1} 
  \begin{tabularx}{0.9\textwidth}{>{\centering\arraybackslash}m{2cm} *{9}{>{\centering\arraybackslash}X}} 
    \toprule
    \multirow{2}{*}{\centering Method} & \multicolumn{3}{c}{CIFAR-10} & \multicolumn{3}{c}{CIFAR-100} & \multicolumn{3}{c}{SVHN} \\
    \cmidrule(lr){2-4} \cmidrule(lr){5-7} \cmidrule(lr){8-10}
    & $0.01$ & $0.05$ & $0.1$ & $0.01$ & $0.05$ & $0.1$ & $0.01$ & $0.05$ & $0.1$ \\
    \midrule
    FedAvg  & 40.30  & 54.23  & 60.08  & 20.27  & 22.29  & 25.86  & 58.81  & 78.84  & 80.55  \\
    FedNova  & 45.18  & 52.99  & 62.91  & 20.30  & 22.24  & 25.74  & 63.53  & 80.24  & 83.26  \\
    FedProx & 40.55  & 54.36  & 59.59  & 20.30  & 22.28  & 25.86  & 58.83  & 78.85  & 80.57  \\
    SCAFFOLD  & 34.71  & 51.28  & 64.23  & 23.55  & 25.48  & 27.19  & 66.16  & 77.66  & 82.44  \\
    FedConcat  & 30.21  & 35.45  & 45.66  & 18.74  & 21.05  & 20.83  & 57.17  & 73.87  & 77.06  \\
    \midrule 
    FedDM  & 61.88  & 63.31  & 63.76  & 31.10  & 32.34  & 34.00  & 80.10  & 82.21  & 83.48  \\
    FedAF & 63.77  & 66.25  & 67.51  & 34.34  & 35.18  & 37.22  & 81.21  & 83.69  & 84.68  \\
    \textbf{FedFD} & \textbf{68.61} & \textbf{70.14} & \textbf{71.54} & \textbf{40.19} & \textbf{41.15} & \textbf{42.62} & \textbf{87.39} & \textbf{87.85} & \textbf{89.14} \\
    \bottomrule
  \end{tabularx}
\end{table*}

\begin{figure*}
    \begin{center}
        \includegraphics[width=1.0\linewidth]{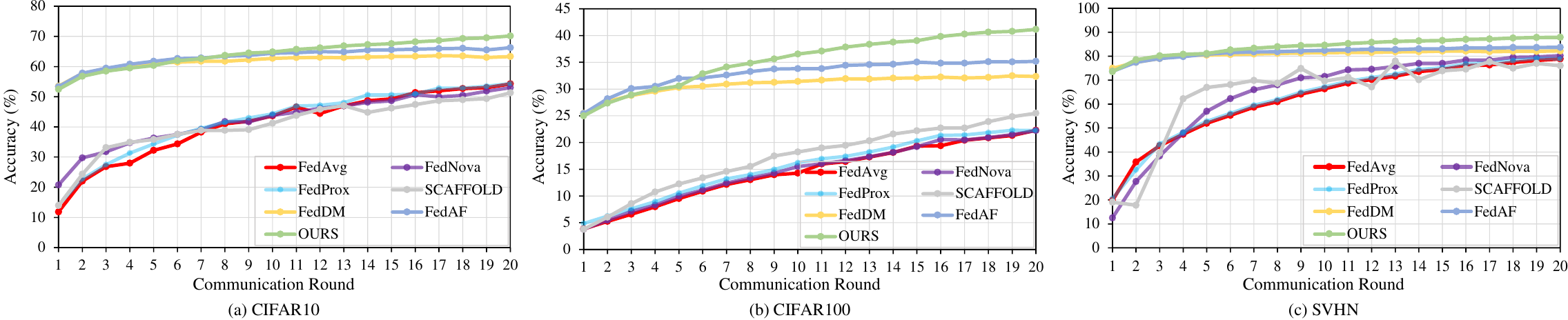}
        \caption{Convergence performance of different FL methods on image datasets.}
        \label{Fig4}
    \end{center}
\end{figure*}

To ensure fairness, the cumulative communication cost for each approach is calculated based on the synthesized data of a single class from the CIFAR-10 \cite{35} dataset. As shown in Fig. \ref{Fig8}, it is evident that regardless of whether the window size $s$ is 8, 12, or 16, the cumulative communication cost is consistently lower than that of entire spatial-domain communication. This further demonstrates the superiority of our approach. Notably, the frequency-domain distribution alignment introduced in the previous section ensures the efficient execution of low-frequency-based curriculum-style communication by reinforcing low-frequency alignment. Furthermore, we compare the security of entire spatial-domain communication and sparse frequency-domain communication in the Theorem 3.

\section{Experiments}
\subsection{Experimental setup}
\textbf{Datasets.} To evaluate the performance of FedFD, we conduct a series of experiments on commonly used datasets across different modalities. For image classification tasks, we select CIFAR-10 \cite{35}, CIFAR-100 \cite{35}, and SVHN \cite{37} datasets. For speech classification tasks, Speech Commands-8 \cite{36} and Speech Commands-10 \cite{36} are used, which are preprocessed into 32×32 size spectrograms. We initialize each class of synthetic data by randomly sampling raw data. When generating synthetic data on $K=10$ clients, the number of images per class (IPC) varies across datasets: initial IPC is set to 10 for CIFAR-10 and SVHN datasets, and to 5 for CIFAR-100, Speech Commands-8, and Speech Commands-10 datasets. The coefficient $m$ is set to the same value as the initial IPC for each dataset. The default and standard train-test splits are used in all experiments.

\textbf{Baselines.} To demonstrate the superiority of FedFD, we compare it with state-of-the-art methods. Aggregation-based FL methods include representative FedAvg \cite{2}, as well as advanced methods FedNova \cite{22}, FedProx \cite{5}, SCAFFOLD \cite{6}, and FedConcat \cite{38}. At the same time, we evaluate FedFD against leading aggregation-free FL methods, specifically FedDM \cite{15} and FedAF \cite{16}. We utilize test accuracy as the prime metric to evaluate the performance of the proposed FedFD method and all baselines \cite{2, 15}.

\begin{table*}[!t]
  \caption{Performance ($\bm{\%}$) comparison of FL methods with different frameworks on Speech Commands-8 dataset and Speech Commands-10 dataset under three levels of heterogeneity ($\bm{\alpha}$).}
  \label{tab2}
  \renewcommand{\arraystretch}{1} 
  \begin{tabularx}{\textwidth}{>{\centering\arraybackslash}m{1.5cm} *{6}{>{\centering\arraybackslash}X}} 
    \toprule
    \multirow{2}{*}{Method} & \multicolumn{3}{c}{Speech Commands-8} & \multicolumn{3}{c}{Speech Commands-10} \\
    \cmidrule(lr){2-4} \cmidrule(lr){5-7}
    & $0.01$ & $0.1$ & $0.2$ & $0.01$ & $0.1$ & $0.2$ \\
    \midrule
    FedAvg    & 55.62 & 64.70 & 75.71 & 40.91 & 68.32 & 71.78 \\
    FedNova  & 51.19 & 63.65 & 76.80 & 25.29 & 69.29 & 71.99 \\
    FedProx  & 55.50 & 64.62 & 75.93 & 41.46 & 68.11 & 71.71 \\
    SCAFFOLD  & 39.14 & 63.41 & 74.45 & 25.82 & 65.82 & 71.36 \\
    FedConcat & 49.71 & 65.79 & 68.72 & 36.69 & 67.90 & 68.49 \\
    \midrule 
    FedDM   & 74.22 & 77.51 & 78.35 & 73.46 & 76.24 & 77.32 \\
    FedAF   & 76.65 & 78.05 & 81.03 & 74.28 & 78.71 & 79.51 \\
    \textbf{FedFD}    & \textbf{81.15} & \textbf{82.71} & \textbf{83.47} & \textbf{81.50} & \textbf{82.82} & \textbf{83.78} \\
    \bottomrule
  \end{tabularx}
\end{table*}

\begin{figure*}
    \begin{center}
        \includegraphics[width=0.7\linewidth]{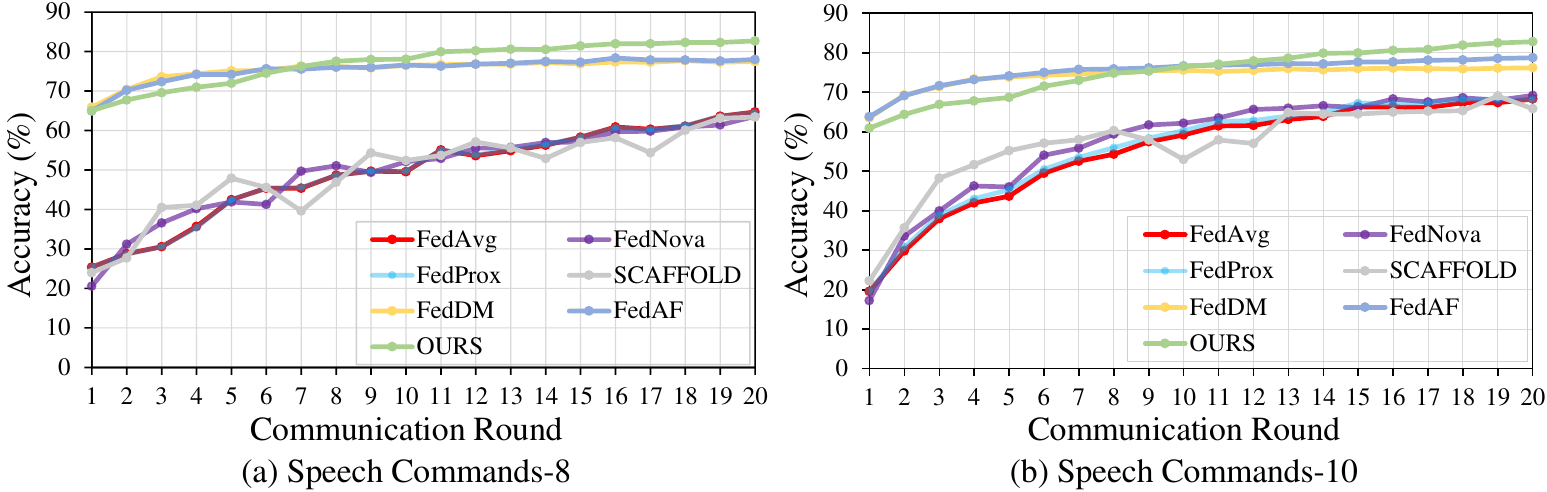}
        \caption{Convergence performance of different FL methods on speech datasets.}
        \label{Fig5}
    \end{center}
\end{figure*}

\textbf{Configuration and hyperparameters.} Following common settings \cite{16}\cite{39}, the training data for each benchmark dataset are divided and distributed to each client based on the Dirichlet distribution, which simulates partitioning of Non-IID data. For image datasets, the Dirichlet coefficient $\alpha$ is set to $\{0.01, 0.05, 0.1\}$, while for speech datasets, $\alpha$ is set to 0.01, 0.1, and 0.2, respectively. Note that a smaller $\alpha$ indicates a higher degree of data heterogeneity. We adopt a local batch size of 64, a learning rate of 1, and 1000 local update steps for local data distillation, while using a batch size of 256 with a learning rate of 0.01 and 500 epochs for global model training. By default, the number of communication rounds is set to 20, and the model architecture is ConvNet \cite{24}. The hyperparameters $\lambda_1$ and $\lambda_2$ are both set to 0.01, which will be discussed in the ablation study section. To avoid bias, we repeat the experiment three times and use the average value as the final result. The proposed FedFD method is implemented in PyTorch on a device equipped with two NVIDIA RTX 3090 GPUs.

\subsection{Comparison to the state-of-the-Art}
\textbf{Overall performance.} The test accuracies of the global model for each method on the image and speech datasets are summarized in Table \ref{tab1} and Table \ref{tab2}, respectively. Obviously, under different conditions, FedFD achieves impressive performance compared to all baselines. Furthermore, aggregation-free baselines generally outperform aggregation-based baselines. When $\alpha = 0.1$ and $0.2$, FedFD exceeds the best aggregation-based baseline, achieving improvements of 7.31\%, 15.43\%, 5.88\%, 6.60\%, and 11.80\% on CIFAR-10, CIFAR-100, SVHN, Speech Commands-8, and Speech Commands-10, respectively. When $\alpha = 0.01$, the performance gap becomes even more significant, with an average performance gain of over 20\%. Figs. \ref{Fig4} and \ref{Fig5} illustrate the test accuracy curves over communication rounds for $\alpha = 0.05$ and $\alpha = 0.1$, respectively. It is evident that aggregation-free methods converge faster than aggregation-based methods, further demonstrating their superiority. 

\begin{figure*}
    \begin{center}
        \includegraphics[width=0.7\linewidth]{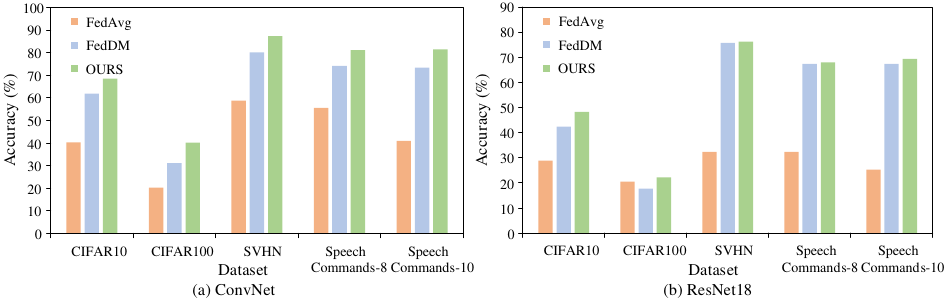}
        \caption{Comparison of performance between FedFD and two representative FL methods using ConvNet and ResNet18 on five datasets. The level of heterogeneity ($\bm{\alpha}$) is set to 0.01.}
        \label{Fig9}
    \end{center}
\end{figure*}

To further demonstrate the advantages of the proposed FedFD method, we employ ResNet18 \cite{40} to compare its performance against two representative algorithms, FedAvg \cite{2} and FedDM \cite{15}, across five datasets. Additionally, to illustrate the generalization performance of FedFD across different network architectures, we also provide performance comparisons using ConvNet \cite{24}. As shown in Fig. \ref{Fig9}, FedFD consistently achieves state-of-the-art performance on both network architectures.

\begin{table*}[!t]
  \caption{Comparison of communication costs between FedFD and two representative FL methods using ConvNet and ResNet18 on five datasets. M and G denote MB and GB, respectively. Speech Commands is abbreviated as SC. The level of heterogeneity ($\bm{\alpha}$) is set to 0.01.}
  \label{tab3}
  \renewcommand{\arraystretch}{1.1} 
  \begin{tabularx}{\textwidth}{>{\centering\arraybackslash}m{3cm} *{6}{>{\centering\arraybackslash}X}}
    \toprule
    \multirow{2}{*}{Dataset} & \multicolumn{3}{c}{ConvNet} & \multicolumn{3}{c}{ResNet18} \\
    \cmidrule(lr){2-4} \cmidrule(lr){5-7}
    & FedAvg & FedDM & \textbf{FedFD} & FedAvg & FedDM & \textbf{FedFD} \\
    \midrule
    CIFAR-10   & 256.0M & 35.2M  & \textbf{21.9M}  & 8.7G  & 35.2M  & \textbf{21.9M} \\
    CIFAR-100  & 403.5M & 326.0M & \textbf{204.0M} & 8.8G  & 326.0M & \textbf{204.0M} \\
    SVHN       & 256.0M & 37.5M  & \textbf{23.4M}  & 8.7G  & 37.5M  & \textbf{23.4M} \\
    SC-8       & 250.9M & 9.4M   & \textbf{5.9M}   & 8.7G  & 9.4M   & \textbf{5.9M} \\
    SC-10      & 254.2M & 12.5M  & \textbf{7.8M}  & 8.7G  & 12.5M  & \textbf{7.8M} \\
    \bottomrule
  \end{tabularx}
\end{table*}

\textbf{Communication costs.} We compare the communication costs of the proposed FedFD method with two representative FL methods across five datasets using different networks. Since the communication costs of aggregation-based baselines are similar, we use FedAvg \cite{2} as a representative. For aggregation-free baselines, both FedDM \cite{15} and FedAF \cite{16} employ entire spatial-domain communication, so we choose FedDM \cite{15} for comparison. To ensure fairness, all network and image sizes are uniformly measured in the float32 format. Table \ref{tab3} summarizes the cumulative communication cost of 20 rounds with 10 clients for each method. As shown in the table, FedAvg \cite{2} incurs relatively high communication costs: up to approximately 403.52MB with ConvNet and around 8.77GB with ResNet18 \cite{40}. In contrast, the communication cost of FedFD is significantly lower, with reductions of up to three orders of magnitude. The communication cost of the aggregation-free baseline represented by FedDM \cite{15} lies between the two methods. Additionally, sending only partial frequency-domain components enhances privacy protection to some extent.

\textbf{Robustness analysis.} As shown in Tables 1 and 2, the proposed FedFD method exhibits outstanding excellence across datasets with different modalities. At the same time, as data heterogeneity increases, the advantage of FedFD becomes more evident. Based on these two points, it exhibits greater robustness compared to the baseline methods. Notably, the above theoretical explanation is strongly supported by the combined overall performance analysis.

\subsection{Ablation study}
\textbf{Impact of core components.} To explore the contributions of the core components in FedFD, we conduct ablation experiments on the CIFAR-10 and SVHN datasets under different $\alpha$ values. The FedDM \cite{15} method is used as the baseline (dubbed BL). On this basis, we introduce the low-frequency-based curriculum-style communication strategy (dubbed LCSC) to demonstrate its advantages. Subsequently, we replace the BL loss with the frequency-domain distribution alignment loss (dubbed FDA) to validate its effect. Finally, we incorporate real data-driven synthetic classification (dubbed RSC) to achieve the frequency-domain-assisted dual-view coordination, which represents the final performance of our FedFD method. The experimental results are summarized in Table \ref{tab4}, with the best results indicated in boldface.

The results in Table \ref{tab4} show that using only BL can achieve commendable performance. The performance is enhanced when the LCSC strategy is used. This shows that LCSC can promote performance improvement while ensuring efficient communication. The performance is further improved when the synthetic data is updated by FDA, which shows that it can promote low-frequency alignment and complement LCSC. With the inclusion of RSC, the performance of FedFD reaches its peak, proving that it can effectively improve the inter-class discrimination of synthetic data. Overall, the performance increases with the increase of components. This shows the importance of each core component and the feasibility of coordination between them.

\begin{figure}
    \begin{center}
        \includegraphics[width=1\linewidth]{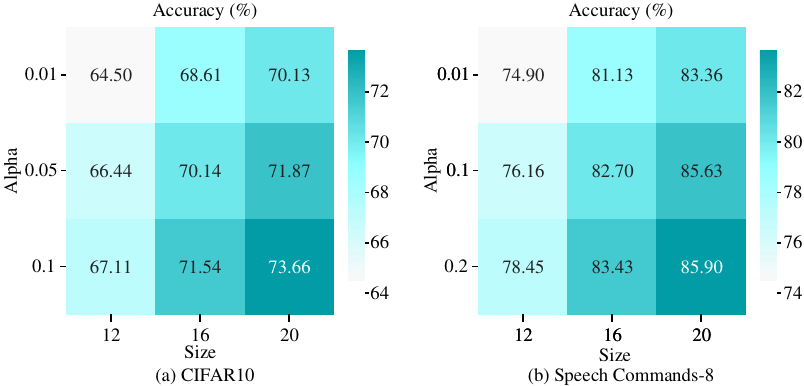}
        \caption{Variation of FedFD with different window sizes $\bm{s}$ on different modality datasets.}
        \label{Fig6}
    \end{center}
\end{figure}

\begin{table*}
  \caption{Performance ($\bm{\%}$) of configurations with different components on CIFAR-10 and SVHN datasets at three levels of heterogeneity ($\bm{\alpha}$). The “+” sign indicates that each component is added on top of the previous configuration.}
  \label{tab4}
  \renewcommand{\arraystretch}{1} 
  \begin{tabularx}{\textwidth}{>{\centering\arraybackslash}m{1.5cm} *{6}{>{\centering\arraybackslash}X}} 
    \toprule
    \multirow{2}{*}{Component} & \multicolumn{3}{c}{CIFAR-10} & \multicolumn{3}{c}{SVHN} \\
    \cmidrule(lr){2-4} \cmidrule(lr){5-7}
    & $0.01$ & $0.05$ & $0.1$ & $0.01$ & $0.05$ & $0.1$ \\
    \midrule
    BL        & 61.88 & 63.31 & 63.76 & 80.10 & 82.21 & 83.48 \\
    +LCSC    & 66.29 & 66.77 & 67.90 & 84.44 & 85.26 & 86.22 \\
    +FDA & 66.78 & 69.35 & 70.97 & 85.49 & 85.69 & 87.21 \\
    \textbf{+RSC} & \textbf{68.61} & \textbf{70.14} & \textbf{71.54} & \textbf{87.39} & \textbf{87.85} & \textbf{89.14} \\
    \bottomrule
  \end{tabularx}
\end{table*}

\textbf{Impact of window size.} When performing frequency domain filtering using binary masks, a larger window size $s$ preserves more information, but it also increases storage requirements. Fig. \ref{Fig6} illustrates the variation in FedFD’s performance with different $s$ values on the CIFAR-10 and Speech Commands-8 datasets. The experimental results indicate that as $s$ increases, the performance of FedFD improves accordingly, which verifies that larger $s$ values can retain more effective information. To balance communication efficiency and performance, $s$ is set to 16 by default in this work.

\textbf{Impact of IPC.} Table \ref{tab5} summarizes the performance variation of FedFD on the CIFAR-10 dataset under different IPC values. It is clear that as IPC increases, the performance of FedFD improves, demonstrating the effectiveness of synthetic data and the potential of FedFD. To balance communication efficiency and overall performance, the default IPC value is set to 10 for the CIFAR-10 dataset. The same approach is used to select the IPC values for other datasets, ensuring consistent performance gains across diverse application scenarios while maintaining an optimal trade-off between resource consumption and data security \cite{49}\cite{16}.

\textbf{Benefit of frequency domain filtering.} When obtaining synthetic data with a window size of $s$=16, four spatial-domain approaches can be adopted: cropping, resizing, random sampling, and gap extraction. To demonstrate that the proposed binary mask-based frequency domain filtering approach can retain more effective information under a limited storage budget, we compare it with the four aforementioned approaches on the CIFAR-10 dataset. As shown in Table \ref{tab6}, under different $\alpha$ values, our approach consistently achieves the best performance, which fully demonstrates its superiority. Moreover, compared to spatial-domain data selection methods, frequency domain filtering further enhances the privacy protection of the original data \cite{16}\cite{19}.

\begin{figure}
    \begin{center}
        \includegraphics[width=0.8\linewidth]{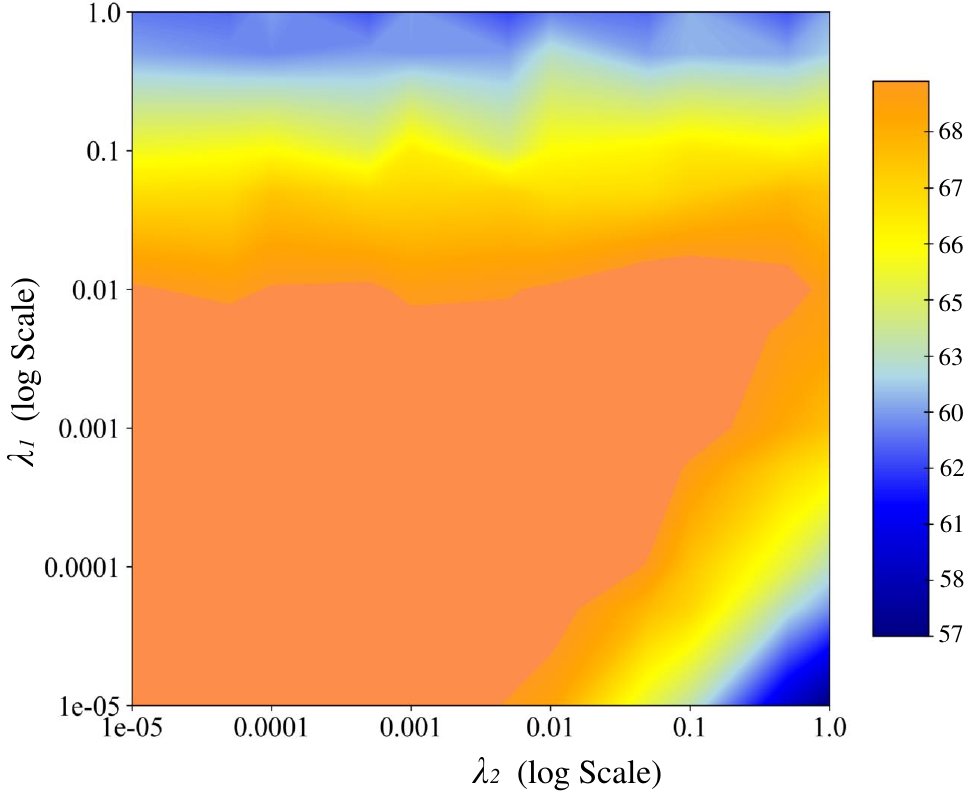}
        \caption{Performance ($\bm{\%}$) grid of FedFD with changing $\lambda_1$ and $\lambda_2$.}
        \label{Fig7}
    \end{center}
\end{figure}

\begin{table}[!t]
  \caption{Performance ($\bm{\%}$) variation of FedFD with different IPC values under three levels of heterogeneity ($\bm{\alpha}$).}
  \label{tab5}
  \renewcommand{\arraystretch}{1} 
  \begin{tabularx}{\columnwidth}{>{\centering\arraybackslash}m{2cm} *{3}{>{\centering\arraybackslash}X}} 
    \toprule
    Coefficient & 0.01 & 0.05 & 0.1 \\
    \midrule
    5      & 63.71 & 66.26 & 67.55 \\
    10    & 68.61 & 70.14 & 71.54 \\
    15 & 70.60 & 73.19 & 73.81 \\
    20 & 72.76 & 73.96 & 75.30 \\
    \bottomrule
  \end{tabularx}
\end{table}

\begin{table}[!t]
  \caption{Effectiveness ($\bm{\%}$) comparison of frequency domain filtering based on binary masks and other data selection approaches.}
  \label{tab6}
  \renewcommand{\arraystretch}{1} 
  \begin{tabularx}{\columnwidth}{>{\centering\arraybackslash}m{2cm} *{3}{>{\centering\arraybackslash}X}} 
    \toprule
    Coefficient & 0.01 & 0.05 & 0.1 \\
    \midrule
    Cropping      & 27.74 & 31.81 & 31.73 \\
    Resizing    & 55.05 & 56.29 & 56.79 \\
    Random & 38.51 & 41.30 & 46.18 \\
    Gap & 62.30 & 63.10 & 65.04 \\
    \midrule
    \textbf{Ours} & \textbf{68.61} & \textbf{70.14} & \textbf{71.54} \\
    \bottomrule
  \end{tabularx}
\end{table}

\textbf{Hyperparameter sensitivity.} The frequency-domain-assisted dual-view coordination expression contains hyperparameters $\lambda_1$ and $\lambda_2$ to balance the contributions of different terms. We perform a grid search \cite{26} on the CIFAR-10 dataset with a Dirichlet coefficient of 0.05, so that readers can choose the appropriate parameter combination according to actual needs. The dark orange area in Fig. \ref{Fig7} indicates the range of parameter values with good performance. Therefore, in all experiments, $\lambda_1$ and $\lambda_2$ are set to 0.01.

\begin{figure*}[!t]
    \begin{center}
        \includegraphics[width=0.8\linewidth]{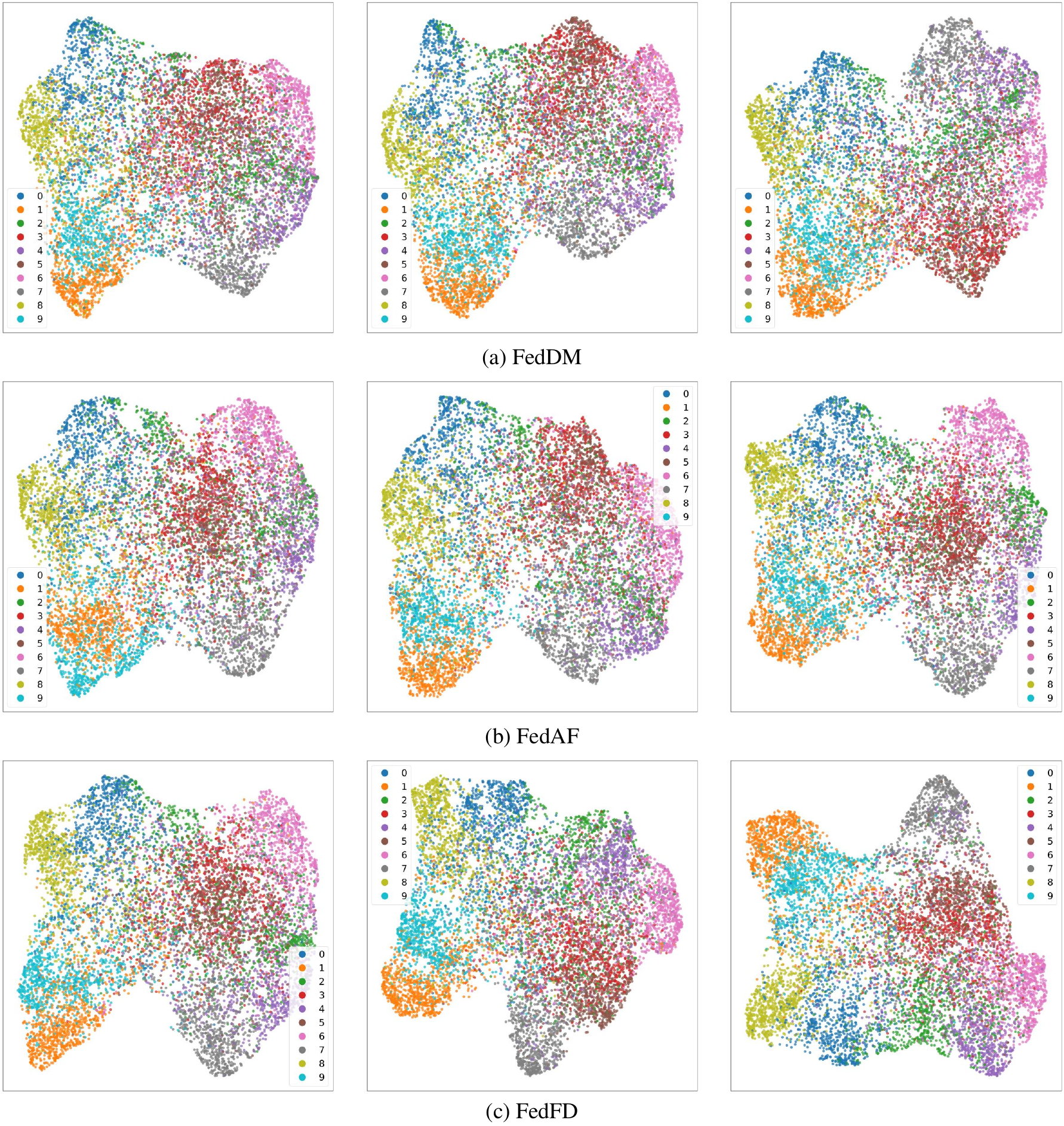}
        \caption{UMAP visualization of embedded features extracted from the global model for three aggregation-free FL methods under three levels of heterogeneity ($\bm{\alpha}$).}
        \label{Fig10}
    \end{center}
\end{figure*}

\begin{figure*}
    \begin{center}
        \includegraphics[width=1.0\linewidth]{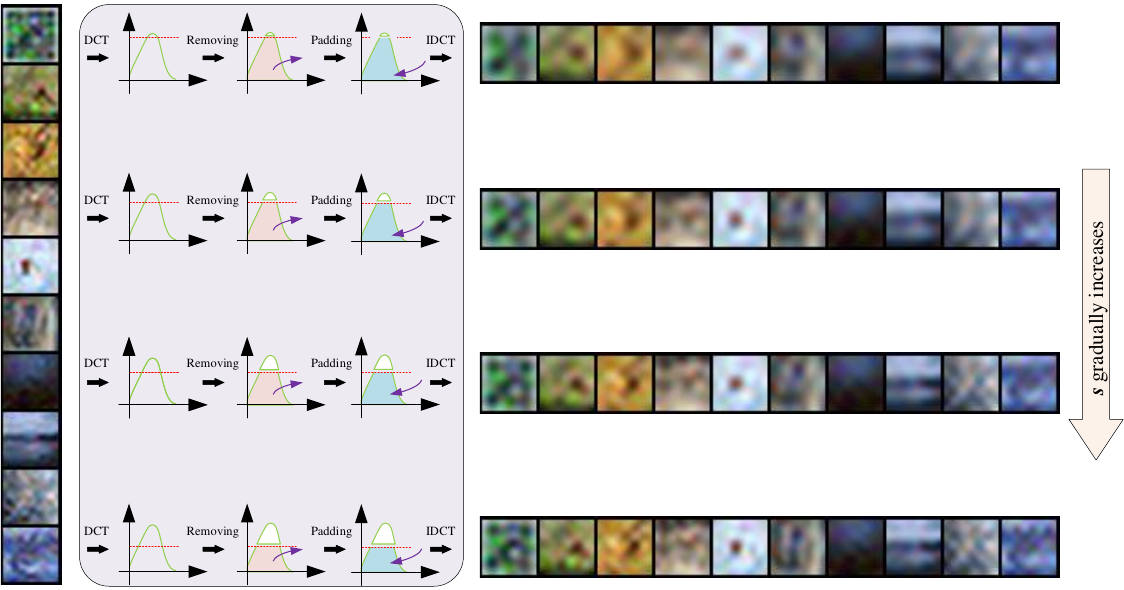}
        \caption{Visualization of synthetic data and the frequency-domain process.}
        \label{Fig11}
    \end{center}
\end{figure*}

\subsection{Visual analysis}
\textbf{Embedded features.}
To clearly demonstrate the advantages of FedFD, we employ UMAP \cite{50} visualization for comparison. Specifically, we evaluate the distribution of embedded features from three aggregation-free FL methods under three different levels of heterogeneity on the CIFAR-10 dataset. As shown in Fig. \ref{Fig10}, it is evident that FedFD consistently achieves the best classification performance, regardless of the level of heterogeneity. This further highlights the superior performance of our method based on qualitative analysis.

\textbf{Synthetic data.}
The role of the frequency-domain-driven dual-view alignment objective in privacy protection is analyzed in detail. To further validate this analysis, we visualize the synthetic data. As shown in Fig. \ref{Fig11}, the frequency-domain filtering process starts by transforming the spatial-domain data into the frequency domain using DCT \cite{29}. Then, higher-frequency components are filtered out using a binary mask, with corresponding positions padded with zeros. Finally, the filtered synthetic data is obtained through IDCT. Meanwhile, from top to bottom, as the value of $s$ increases, more information is preserved in the frequency domain. The left part of Fig. \ref{Fig11} shows that the synthetic data generated on each client already serves the purpose of privacy protection, and after applying frequency domain filtering, this protection is further enhanced.

\section{Conclusion}
In this paper, we present a novel FedFD method that leverages the potential of the frequency domain in FL to balance communication efficiency and performance. Unlike existing aggregation-free FL methods, FedFD explores frequency-domain-assisted dual-view coordination to match the frequency-domain distribution, while emphasizing inter-class discrimination. This process effectively enhances the quality of the low-frequency components in the synthetic data. Accordingly, a low-frequency-based curriculum-style communication strategy is designed to send low-frequency components with higher energy concentration within a limited storage budget, which improves efficiency while further boosting performance. Extensive experiments on multiple datasets with different modalities demonstrate that the proposed FedFD method achieves the state-of-the-art.

\printcredits

\bibliographystyle{cas-model2-names}

\bibliography{cas-refs}
\end{document}